# Cardiac fat segmentation using computed tomography and an image-to-image conditional generative adversarial neural network

Guilherme Santos da Silva [a], Dalcimar Casanova [a], Jefferson Tales Oliva [a], Erick Oliveira Rodrigues [a,b,*]

[a] Academic Department of Informatics, Universidade Tecnológica Federal do Paraná (UTFPR), Pato Branco, 85503-390, Brazil
[b] Graduate Program of Production and Systems Engineering, Universidade Tecnológica Federal do Paraná (UTFPR), Pato Branco, 85503-390, Brazil

ARTICLE INFO

ABSTRACT

In recent years, research has highlighted the association between increased adipose tissue surrounding the human heart and elevated susceptibility to cardiovascular diseases such as atrial fibrillation and coronary heart disease. However, the manual segmentation of these fat deposits has not been widely implemented in clinical practice due to the substantial workload it entails for medical professionals and the associated costs. Consequently, the demand for more precise and time-efficient quantitative analysis has driven the emergence of novel computational methods for fat segmentation. This study presents a novel deep learning-based methodology that offers autonomous segmentation and quantification of two distinct types of cardiac fat deposits. The proposed approach leverages the *pix2pix* network, a generative conditional adversarial network primarily designed for image-to-image translation tasks. By applying this network architecture, we aim to investigate its efficacy in tackling the specific challenge of cardiac fat segmentation, despite not being originally tailored for this purpose. The two types of fat deposits of interest in this study are referred to as epicardial and mediastinal fats, which are spatially separated by the pericardium. The experimental results demonstrated an average accuracy of 99.08% and f1-score 98.73 for the segmentation of the epicardial fat and 97.90% of accuracy and f1-score of 98.40 for the mediastinal fat. These findings represent the high precision and overlap agreement achieved by the proposed methodology. In comparison to existing studies, our approach exhibited superior performance in terms of f1-score and run time, enabling the images to be segmented in real time.

## 1. Introduction

In recent years, studies have shown [1–7] that the increase in the amount of fat surrounding the heart was associated with a higher risk of triggering some cardiovascular diseases, such as atrial fibrillation (cardiac arrhythmia that can cause a stroke) and coronary heart disease (obstruction of the arteries of the heart that can lead to acute myocardial infarction) [1]. Currently, some image acquisition modalities are adequate for the quantification of these adipose tissues such as magnetic resonance imaging (MRI), echocardiography and computed tomography (CT). Although these modalities have been widely used in several studies in the literature, computed tomography provides a more accurate assessment of adipose tissues due to its greater spatial resolution when compared to ultrasound and magnetic resonance imaging. In addition, CT is also widely used to compute the coronary calcium score [1].

CT creates detailed images of most varied tissues of the human body. The need to more precisely perform quantitative analysis on these images has driven the development of new computational methods for organ segmentation [8]. Image segmentation refers to the process of classifying each pixel of an image into a certain category. Therefore, when the image is segmented, processes such as quantification, identification and interpretation of the content is facilitated.

Several works [9–12,7,13] emerged after the introduction of semantic segmentation, which is the task of classifying each one of the pixels in an image into classes. Rodrigues et al. [1] employed a methodology to automatically segment and consequently discriminate epicardial and mediastinal fats in cardiac CT images using classic machine learning techniques (features extraction, parameter optimization and classifier comparison) [14].

Along with the growth of machine and deep learning [12,14], imaging [1,15] and optimization techniques [16], new possibilities emerged,

* Corresponding author at: Academic Department of Informatics, Universidade Tecnológica Federal do Paraná (UTFPR), Pato Branco, 85503-390, Brazil.
*E-mail address:* erickrodrigues@utfpr.edu.br (E.O. Rodrigues).

for example, algorithms specifically focused on image segmentation and more sophisticated neural networks, such as U-net [17], which is a convolutional neural network developed for biomedical image segmentation, and pix2pix, a generative network dedicated to the task of image-to-image translation. Most works in the literature that perform cardiac fat segmentation is outdated in terms of technology. In this work, we try a novel method for this type of task that has not been tested before [18–21]. We also aim to improve the run time that is required to segment an image.

When introducing the pix2pix network, Isola et al. [22] found that this conditional adversarial generative networks (cGANs) are able to solve the semantic segmentation problems moderately. Although cGANs achieve some success, they are not the best methods available to solve to be used with semantic segmentation [22], or at least there were not primarily designed for this. Nevertheless, we illustrate that this methodology yields very satisfactory outcomes in the segmentation of cardiac fat.

This work proposes a new method for the cardiac fat segmentation using the pix2pix network, a conditional adversary generative network. This network was ideally created to perform image-to-image translation. In other words, it takes an input image and generates an output image according to the training applied to the network. The aim of this work is to evaluate the performance of this model in terms of semantic segmentation. We compare the approach in terms of accuracy and run time to models widely accepted and used as reference in the literature, such as the U-net network [23].

## 2. Materials and methods

Numerous studies have established associations between the quantity of epicardial fat and the progression of coronary artery calcification [1]. For example, Hoffmann et al. [24] discovered that patients with high-risk coronary lesions exhibited nearly double the epicardial fat volume compared to those without coronary artery calcification. Additionally, the volume of epicardial adipose tissue has been linked to various cardiovascular risk factors and outcomes. These include diastolic filling abnormalities, myocardial infarction, atrial fibrillation and outcomes of ablation procedures, carotid stiffness, atherosclerosis, and numerous other factors [1]. Moreover, Chen et al. [25] demonstrated that a high coronary artery calcium score is associated with a higher general incidence of cancer.

In this work, we replicate the definition proposed by [1] in order to establish clear and consistent terminology for cardiac fats. We define the fat within the epicardium as “epicardial,” in alignment with the prevailing terminology in published literature. Similarly, following the same logical model of considering the “first outer anatomical container,” we determine that the appropriate term for the fat located on the external surface of the heart or fibrous pericardium is “mediastinal.” In other words, the mediastinal fat corresponds to the area within the anterior, middle, and posterior mediastinum. It should be emphasized that the mediastinal fat is defined as such as long as it is not located within the epicardium, i.e., it does not reside within the epicardium.

Shifting the attention to the imaging technique, computed tomography has two important advantages over conventional radiographs [1]: three-dimensional image reconstructions and the ability to quantify X-ray attenuation. Attenuation is expressed in CT using the Hounsfield Unit Scale (HU). X-ray beams used for diagnostic radiology are composed of photons with a broad spectrum of energies. Each system and manufacturer incorporates a unique combination of the X-ray source, detector array and projection geometry.

In order to read the appropriate data corresponding to the cardiac adipose tissue in CT scans we need to consider an interval close to −100 HU. This corresponds to the total human body fat, as shown in Table 1. In this work, the image database available in [26] is used, which covers a range from −200 HU to −30 HU. This chosen range encompasses the HU range referring to the adipose tissue, it fits properly in an 8-bit deep

**Table 1**
Hounsfield scale.

| Substance | HU |
|---|---|
| Air | -1000 |
| Lung | -500 |
| Fat | -100 to -50 |
| Water | 0 |
| Cerebrospinal fluid | 15 |
| Kidney | 30 |
| Blood | 30 to 45 |
| Muscle | 10 to 40 |
| Gray matter | 37 to 45 |
| White mass | 20 to 30 |
| Liver | 40 to 60 |
| Soft Tissues, Contrast | 100 to 300 |
| Bone | 700 to 3000 |

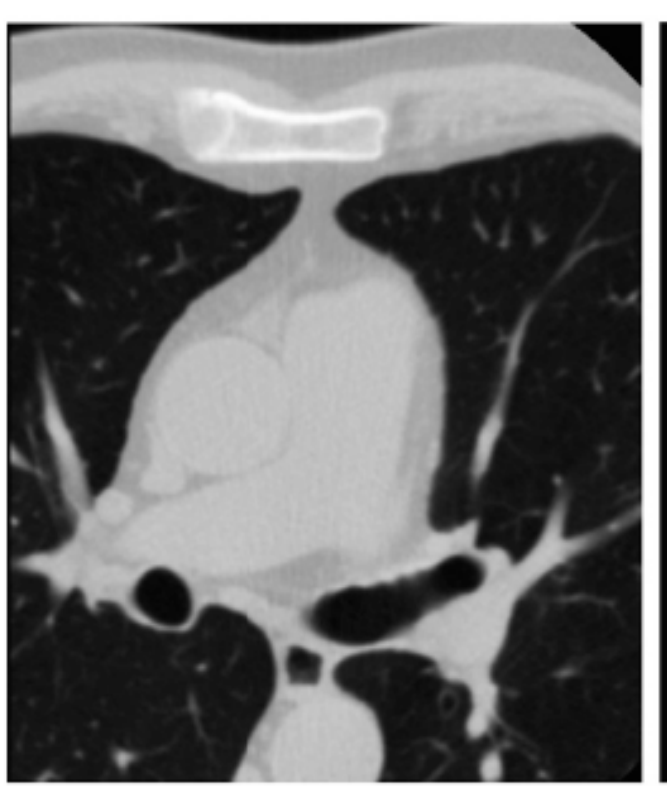
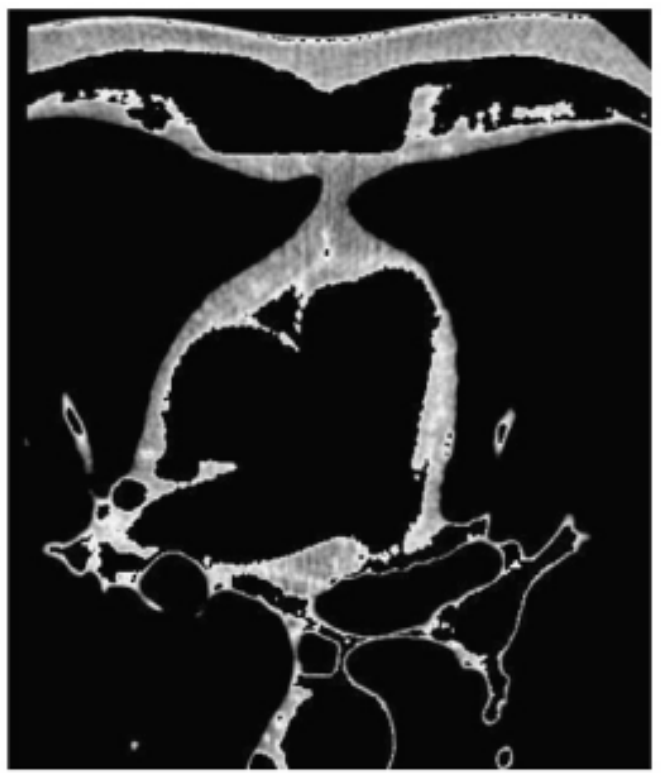

**Fig. 1.** CT image with a range of -1000 to 720 HU (left) and the same CT image with a range of -200 to 30 HU (right).

image and no interpolation is necessary, avoiding any potential loss of data.

Fig. 1 illustrates on the left a computed tomography image of a human heart showing the range from -1000 to 720 HU, encompassing tissues, fluids, air and fats. The image on the right is the same slice presented on the left but it shows just the range used in this work: -200 to -30 HU. The background is represented by the pure black color (intensity = 0).

In Fig. 2 we show some example images. The images on the left are the input using the -200 to 30 HU range and the right is their respective ground truth, annotated by hand by two specialists. More information about the database can be found at [1].

This database is composed of resources extracted from twenty patients randomly chosen at Hospital Universitário Clementino Fraga, located in Rio de Janeiro, Brazil. The images were manually segmented by a physician and a computer scientist and consist of segmentation of epicardial and mediastinal fats in non-contrast cardiac computed tomography scans, totaling 843 images. The red color represents epicardial fat, green color represents mediastinal fat, and the blue color represents the gap between epicardial and mediastinal fat, which can also be interpreted as pericardium [26].

The red (epicardial fat) is comprised within the concentric sac called epicardium, “whithin” the human heart. The green (mediastinal fat) correspond to fats within the mediastinum, where these fats can be directly fixed to the outer surface of the human heart. The epicardial fat is the most important fat in terms of cardiac risk factors, but the mediastinal fat is also important and both correlate to each other, as shown in previous work [27]. The greater the epicardial fat the greater the mediastinal fat and the contrary also applies.

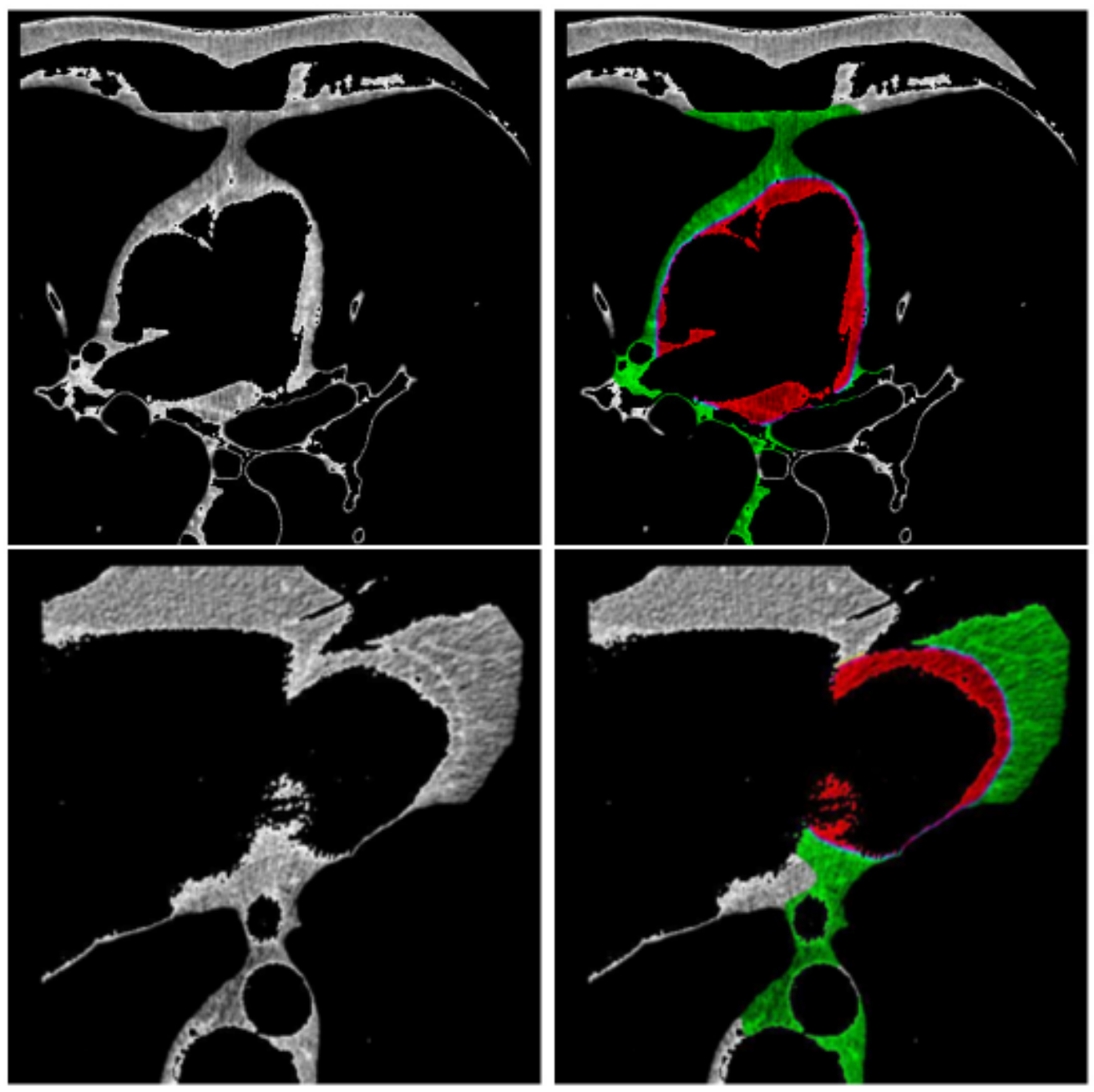

**Fig. 2.** Input image (left) and target image (right).

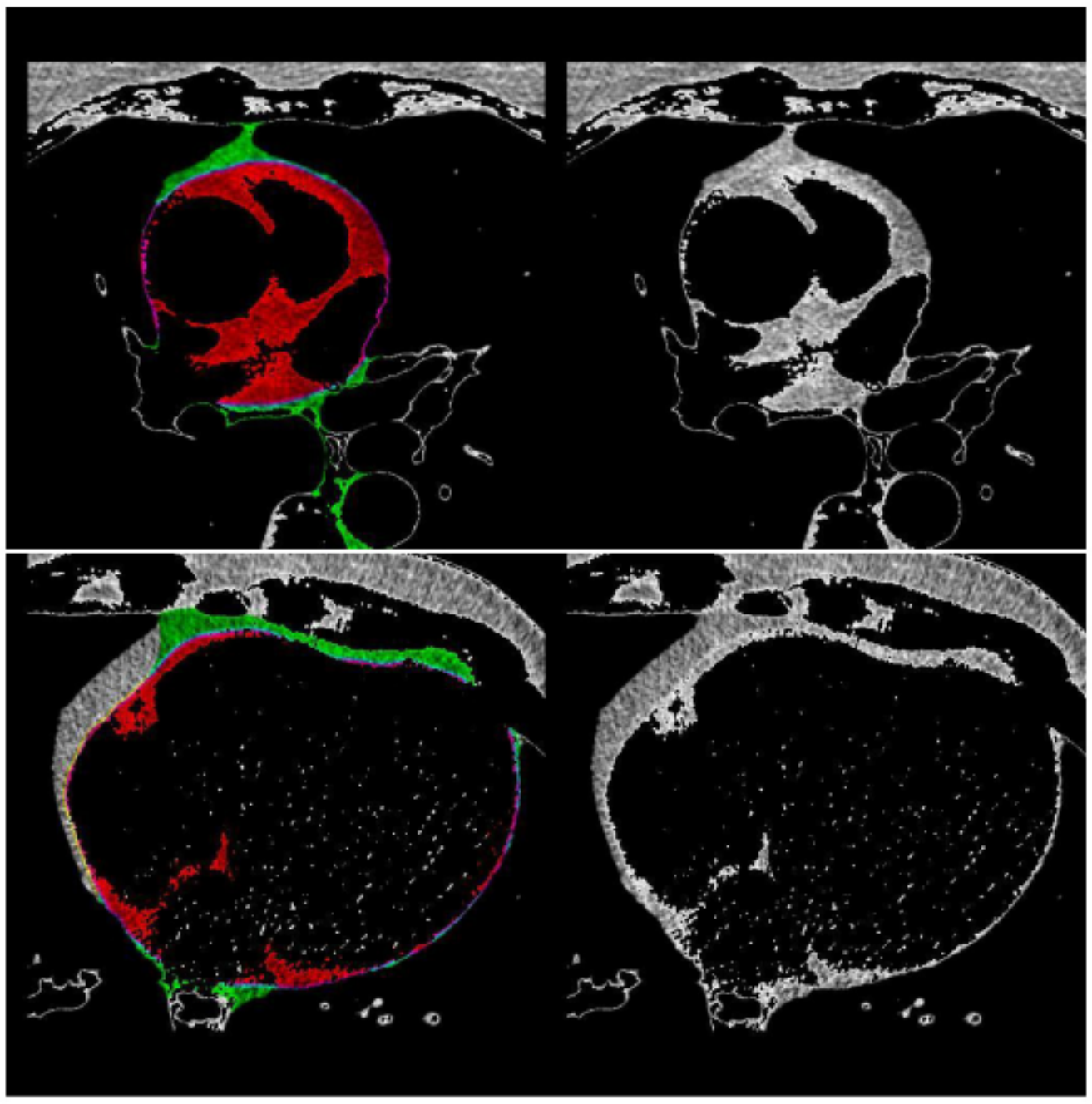

**Fig. 3.** Experiment with input image resizing. Image pair {B,A}.

**Table 2**
Allocation of images in datasets for training, validation and test steps.

| Dataset | Number of images | Size (pixels) |
|---|---|---|
| Training | 562 | 512 x 512 |
| Validation | 169 | 512 x 512 |
| Test | 112 | 512 x 512 |

## 2.1. *Experiments*

For the experiments below, the images were randomly ordered to ensure greater diversity in training and, consequently, a better result. According to the recommendation of Isola et al. [22], the database was separated using the 3-way holdout method, which consists of three sets, one for training, one for validation and one for testing, as shown in Table 2. A set of validation data is needed to evaluate the model during training. This dataset can be used to find and optimize the best model [28] in order to reduce overfitting and fine-tune hyperparameters.

The selection of the best segmentation model is done through the variation of hyperparameters, such as the variation in the learning rate, number of epochs and batch size. During training, some processes can be performed, such as flipping an image horizontally or resizing it to a larger size and then to its original size (parameters “flip” and “resize and crop”, respectively), to generate more data for training, this being the method of data augmentation, in which the idea is to increase the amount and diversity of data in your dataset.

### 2.1.1. *Experiment resizing the input image*

The initial experiment is conducted with the objective of evaluating the performance of the pix2pix model within the context of a supply problem. This evaluation refrains from employing any pre-processing or post-processing methodologies aimed at result enhancement. It is important to emphasize that the main function of the pix2pix model is not centered on explicit pixel classification, but rather on the generation of a new image that closely resembles the ground truth.

Prior to enabling the pix2pix network to acquire proficiency in cardiac adipose tissue segmentation, a requisite adjustment in data formatting is imperative. This necessity arises from the demand for paired data as inputs. Consequently, the procedure involves juxtaposing an input image denoted as A alongside its corresponding output image termed B, which inherently constitutes the segmented form of image A. The combination of these images is executed to create a singular composite representation.

It is noteworthy that images forming a matched pair require identical dimensions. To automate this, a Python script was developed, yielding outcomes exemplified in Fig. 3, wherein an image pair, denoted as B, A, is exhibited. This orchestrated alignment subsequently facilitates the network capacity to effectuate translations between A and B, or conversely, between B and A. It should be emphasized that in the present experimental configuration, the input images undergo resizing processes that result in dimensions smaller than their original proportions, thereby leading to the inevitable loss of certain pixels.

### 2.1.2. *Experiment using just the epicardial fat*

The second experiment involved a distinct methodology wherein the isolation of cardiac adipose tissue was undertaken within the input images, as visually depicted in Fig. 4. This approach entailed the individualized training of the segmented cardiac fat components, thereby rendering them amenable to binary image treatment. Specifically, within a matrix of dimensions NxN, with N representing the image size, pixels devoid of adipose tissue were assigned a numerical value of 0, while pixels encompassing adipose tissue were allocated a value of 1. It is crucial to underline that this experimental configuration exclusively focused on the epicardial fat fraction, denoted by the red regions, which is usually the most important.

Fig. 5 portrays a pair of images, featuring a segmented representation on the left and its corresponding ground truth counterpart on the right. Notably, the output yielded by the conditional Generative Adversarial Network (cGAN) demonstrates a remarkable proximity to the manually executed segmentation. Nonetheless, certain imperfections manifest in the form of minute “holes” dispersed across the adipose tissue region. These voids correspond to background pixels, as visually highlighted in Fig. 6. The presence of such diminutive “holes” can potentially lead to a diminution in the overall performance of the model.

The utilization of mathematical morphology [29] was also explored to address the presence of “holes” within the images. In this regard, the applied morphological post-processing technique was that of **closing**, encompassing a dilation [29] operation succeeded by an erosion operation on the resultant. Overall, the goal is to re-establish connections of the red color, indicative of epicardial fat, while preserving the fundamental dimensions and contours of the initial segmentation. Additionally, this operation serves to enhance the smoothness of the outer boundaries.

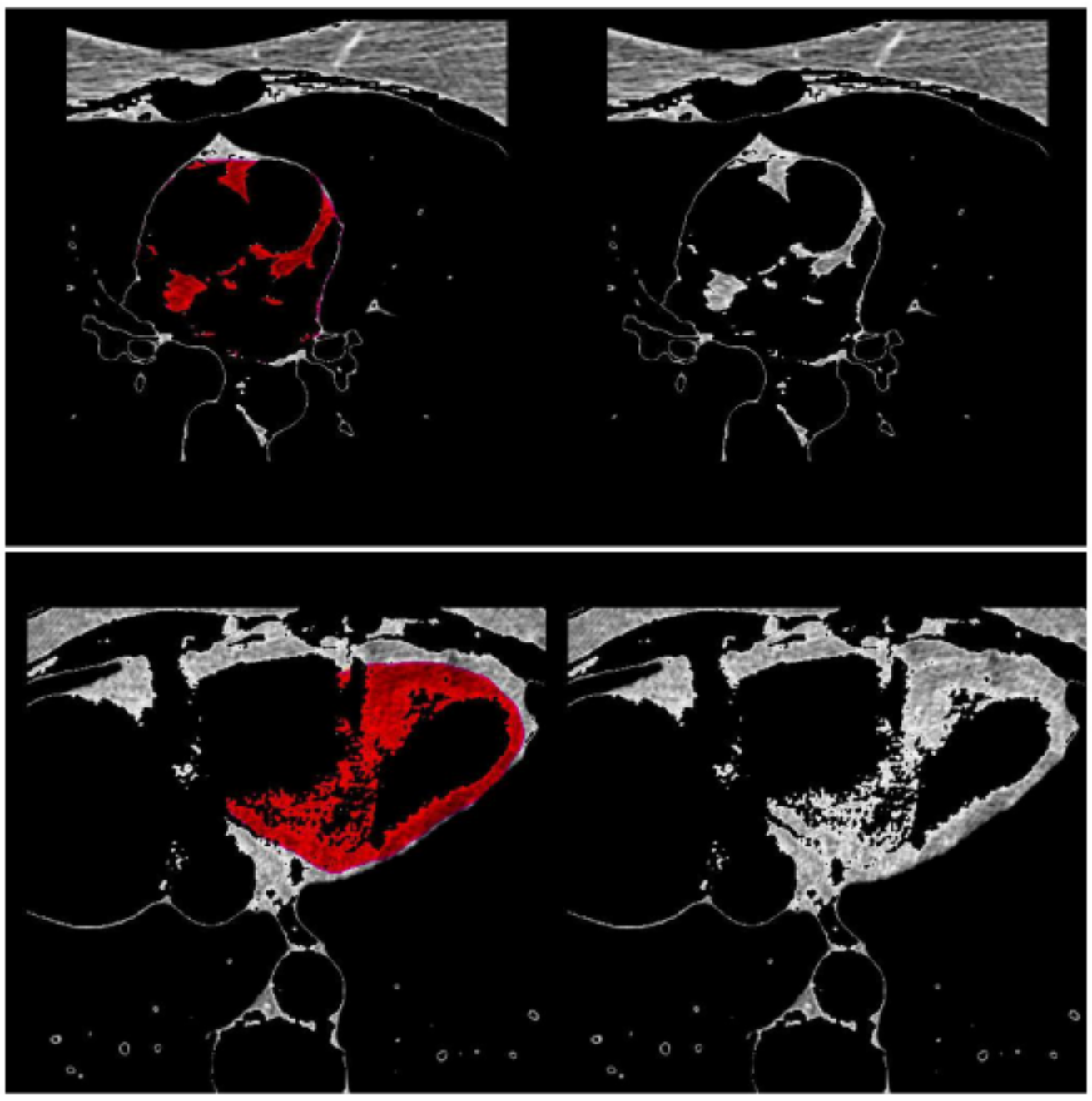

**Fig. 4.** Experiment with one fat. Image pair {B,A}.

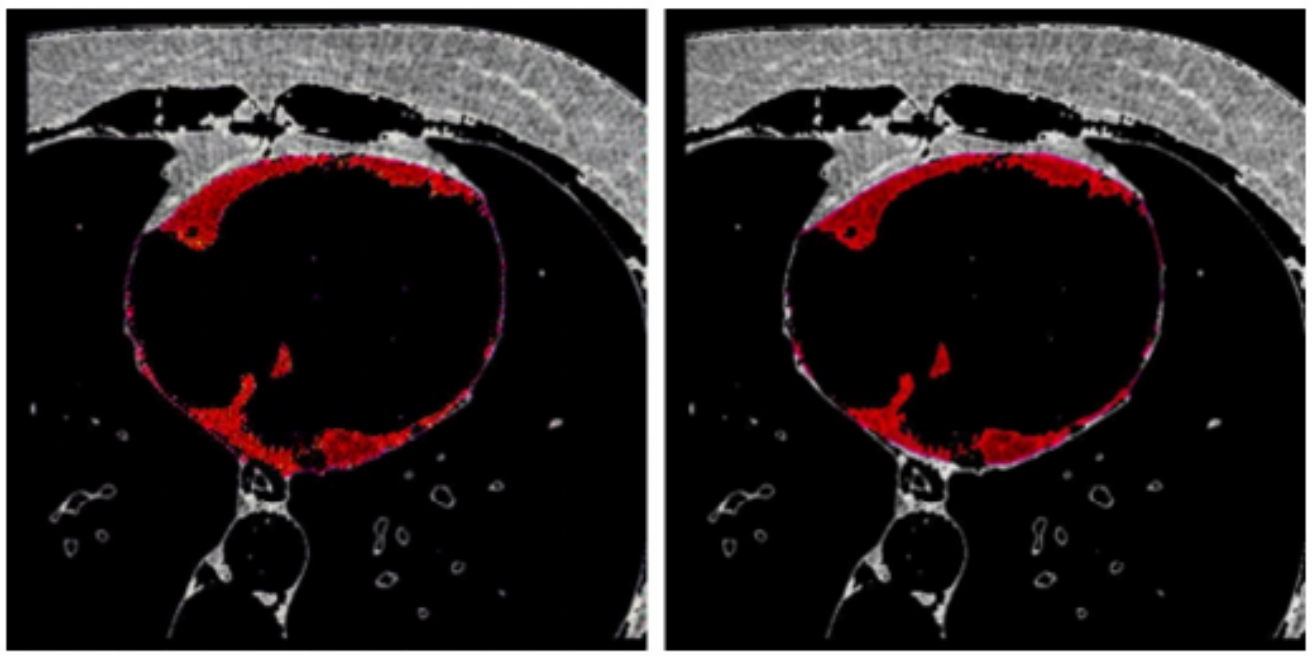

**Fig. 5.** Experiment with just the pericardial fat (one fat). Comparison of a segmented image (left) and ground truth image (right).

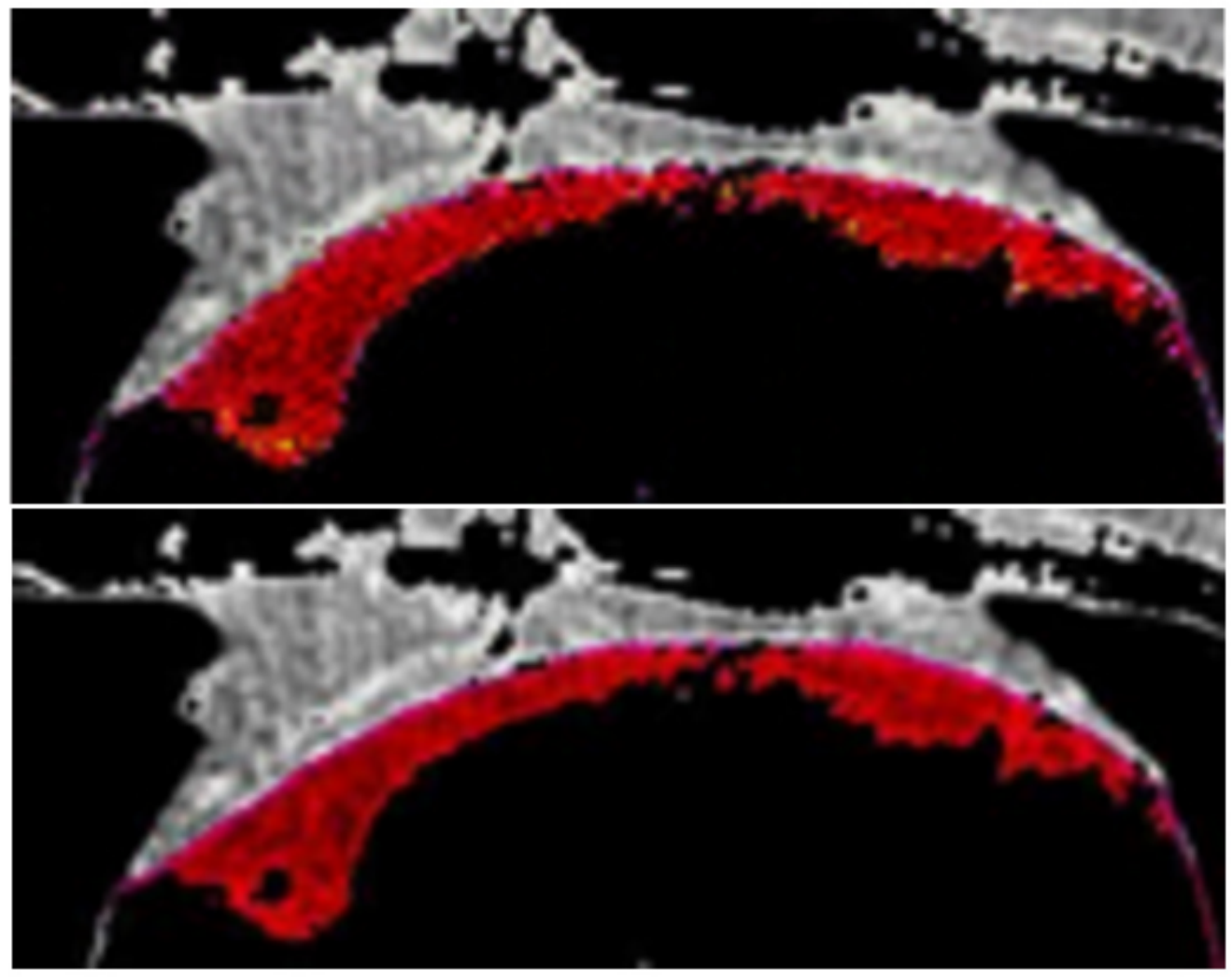

**Fig. 6.** Experiment with just one fat. In the segmented image above, small "holes" are discernible in comparison to the ground truth image at the bottom.

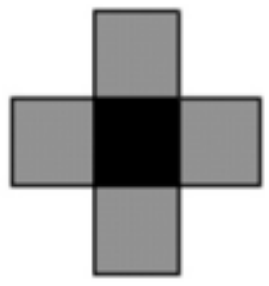

**Fig. 7.** Basic structuring element with a cross shape.

The morphological operations of dilation and later erosion were performed using the binary image while adopting a cross-shaped structuring element shown in Fig. 7, as the image size is relatively small (256x256 pixels).

The outcomes of these operations are showcased in Fig. 8. In this presentation, the top-left image depicts the state prior to the implementation of the morphological operators. Subsequently, the top-right image portrays the condition following the dilation procedure, while the image at the bottom reflects the outcome post the erosion operation.

When traversing the resulting binary image pixel by pixel and finding RGB values corresponding to the white color, the original CT image had the pixel of this same position colored red, classifying it as epicardial fat, as seen in Fig. 9, where the image on the top left is the binary image resulting from the closing process, the image in the top right is the cardiac fat segmentation with the post-processing and the image at the bottom is the ground truth.

### *2.1.3. Experiment using the original image size*

Processes involving image resizing encompass certain drawbacks, notably the loss of data. This deficiency becomes evident when an image undergoes reduction in dimensions. For instance, when transitioning from an original resolution of 1024x1024 pixels to 512x512 pixels, the resulting image does not retain the entirety of the initial pixel data.

In this sense, the objective of the third experiment is to address this aspect. The input images are initially sized at 512x512 pixels. However, the *pix2pix* network, as a standard practice during the initial phases of training, subjects these images to resizing procedures, ultimately attaining dimensions of 256x256 pixels, a prevalent size employed in conjunction with this conditional Generative Adversarial Network (cGAN). Fig. 10 serves to illustrate some representative input images within the purview of this particular experiment.

Therefore, for this experiment, we forced the network to work with the original image resolution (512x512), using this image directly as input. However, results show that it is slower to train and can be less accurate.

In the context of this scenario, the network parameters were specified with the values 572 and 512, allocated respectively to *load_size* and *crop_size*. These selections were made to align with the initial dimensions of the input images and to maintain proportionality in relation to the default values inherent to *pix2pix*, which are 286 and 256, correspondingly.

### *2.1.4. Experiment with 4 image patches*

The last experiment seeks a second approach to address the problem of information loss. As a solution, each input image of size 512x512 pixels can be divided into 4 images of size 256x256, as shown in Fig. 11.

This strategy culminates in a quadrupling of the dataset volume, subsequently engendering augmented durations for both training and segmentation procedures. Following the completion of the concatenation process for the matching images, the input images designated for this experiment are thereby generated. A single image patch pertaining to this specific experiment is showcased in Fig. 12.

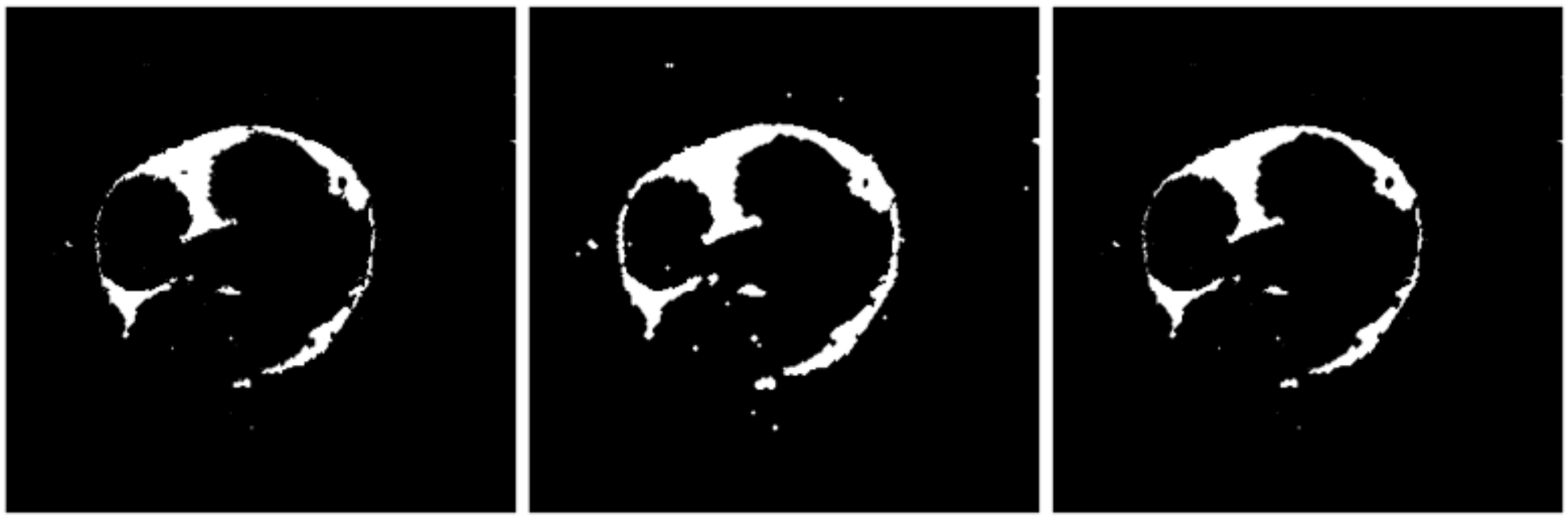

**Fig. 8.** Experiment with just the epicardial fat (one fat). Closing operation (removing the holes).

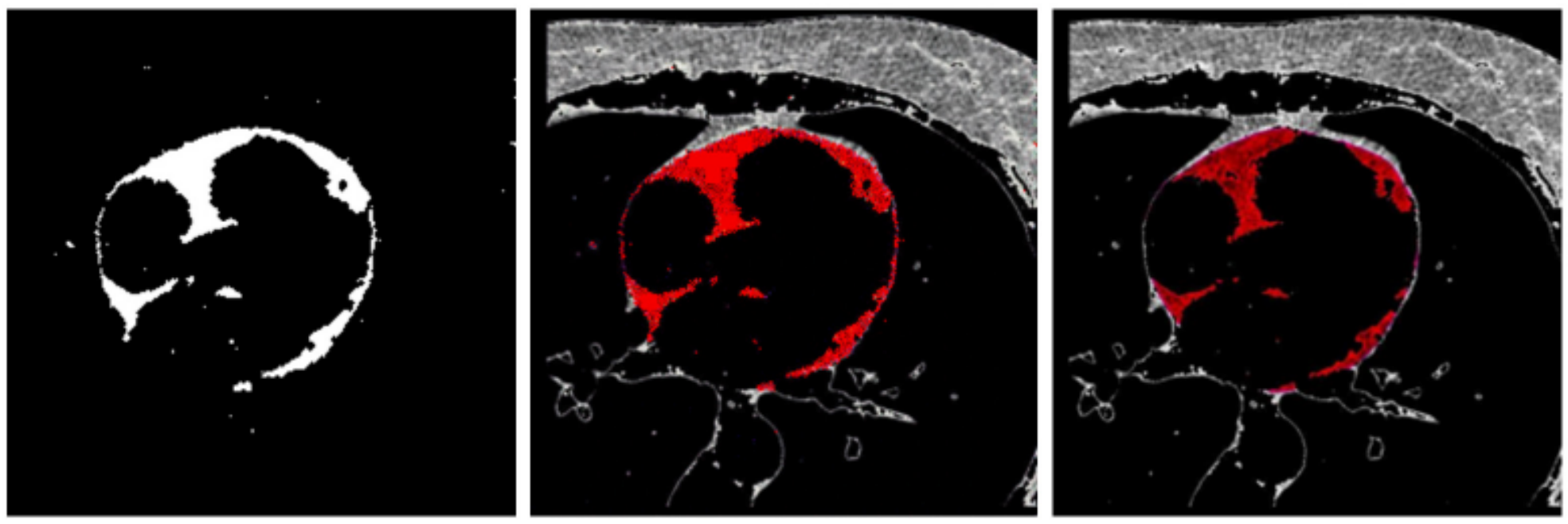

**Fig. 9.** Experiment with just the epicardial fat (one fat).

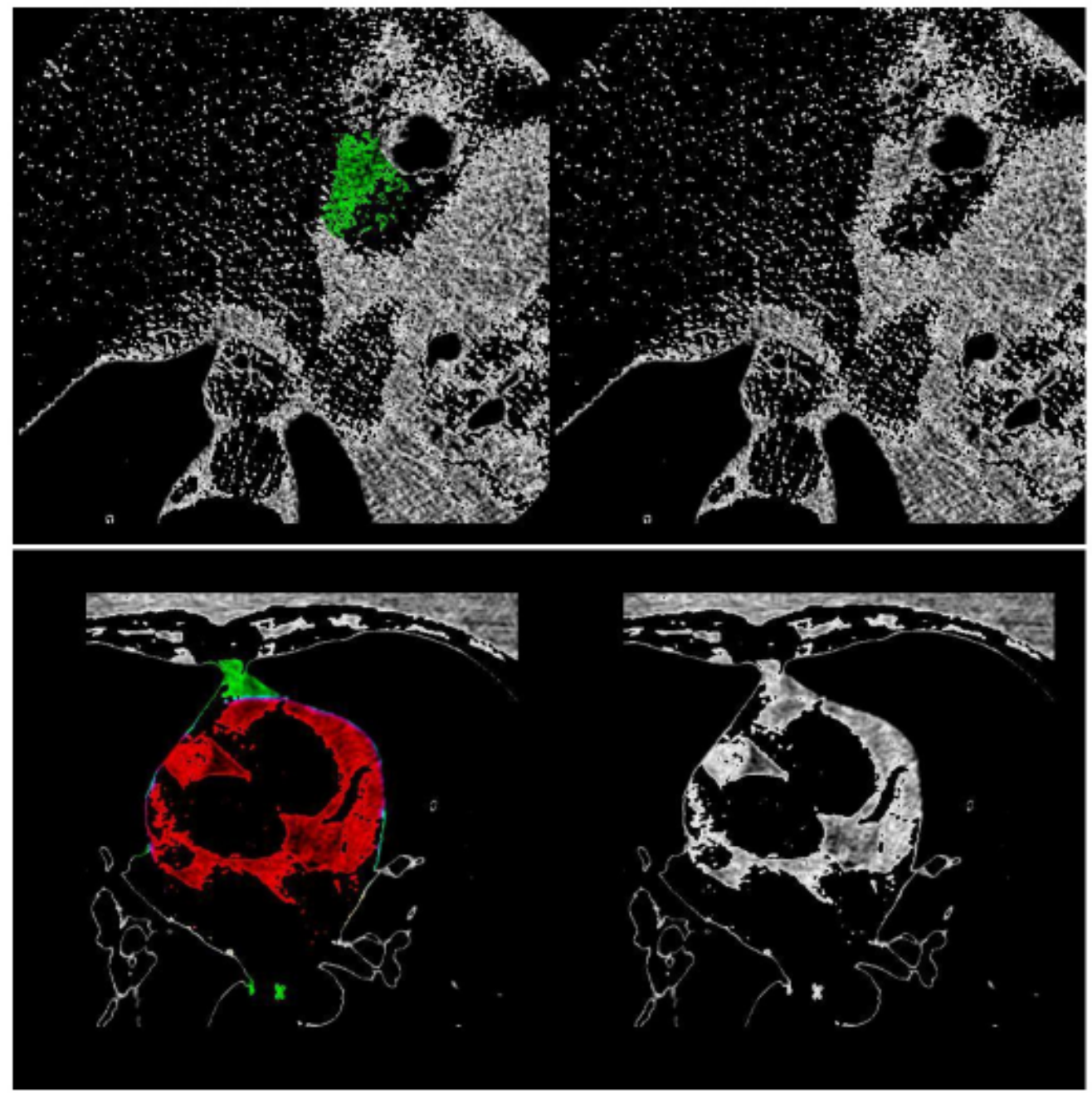

**Fig. 10.** Experiment with original size of the input image. Image pair {B,A}.

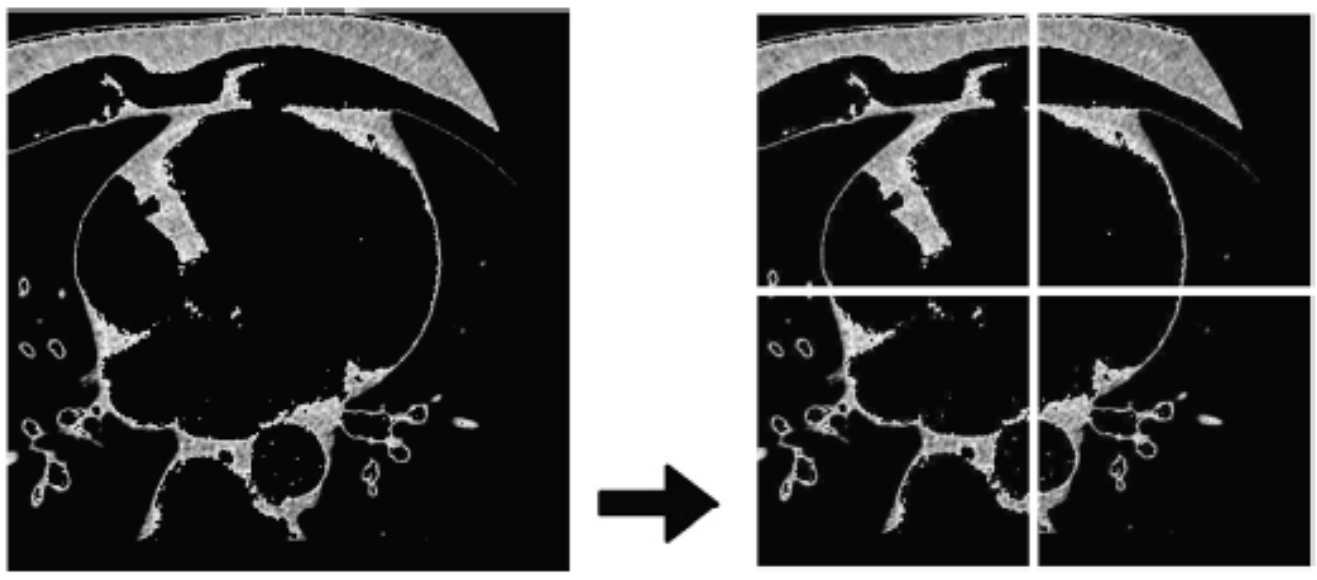

**Fig. 11.** Experiment with the patches. Splitting the input image into 4 patches.

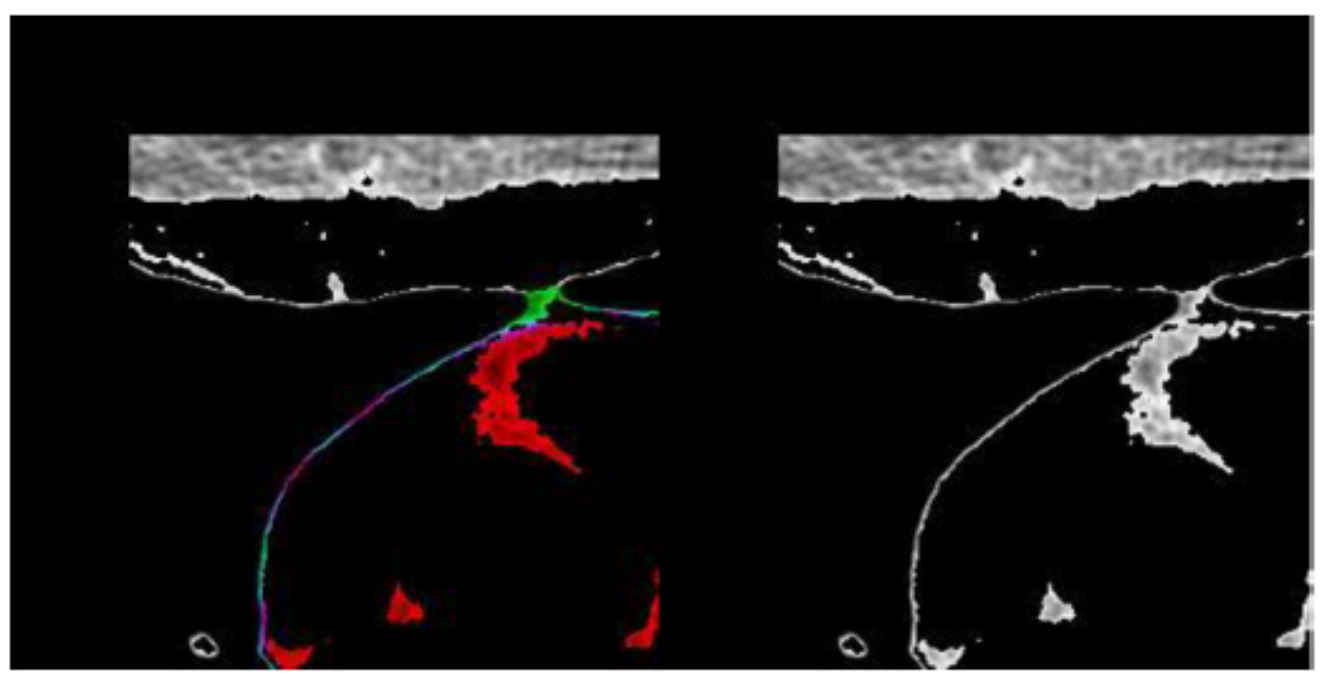

**Fig. 12.** Experiment with the image patches. Image pair {B,A}.

### *2.2. Evaluation measures*

While established as conventional performance metrics, the True Positive (TP), True Negative (TN), False Positive (FP), False Negative (FN), and accuracy rates, although relevant, can exhibit limitations in effectively capturing segmentation efficiency. Notably, these metrics do not adeptly address the presence of class imbalance, a factor that may not be readily discernible.

In cases of pronounced class imbalance, specific classes or categories tend to dominate an image, while others represent only a minor fraction. The background pixels in this case occupy a significant portion of the image. Consequently, the aforementioned metrics might yield favorable outcomes due to accurate targeting of background pixels, while potentially not effectively optimizing for the fat pixels, leading to imbalanced performance.

In order to improve the comparison in this sense, we calculate the $F_1$-score and the IoU, which are measures that represent the similarity between two data sets as a single value between 0 and 1, and can be expressed also in terms of TP, FP and FN, as presented in Equations (1) and (2). These measurements are commonly used to quantify the performance of image segmentation methods.

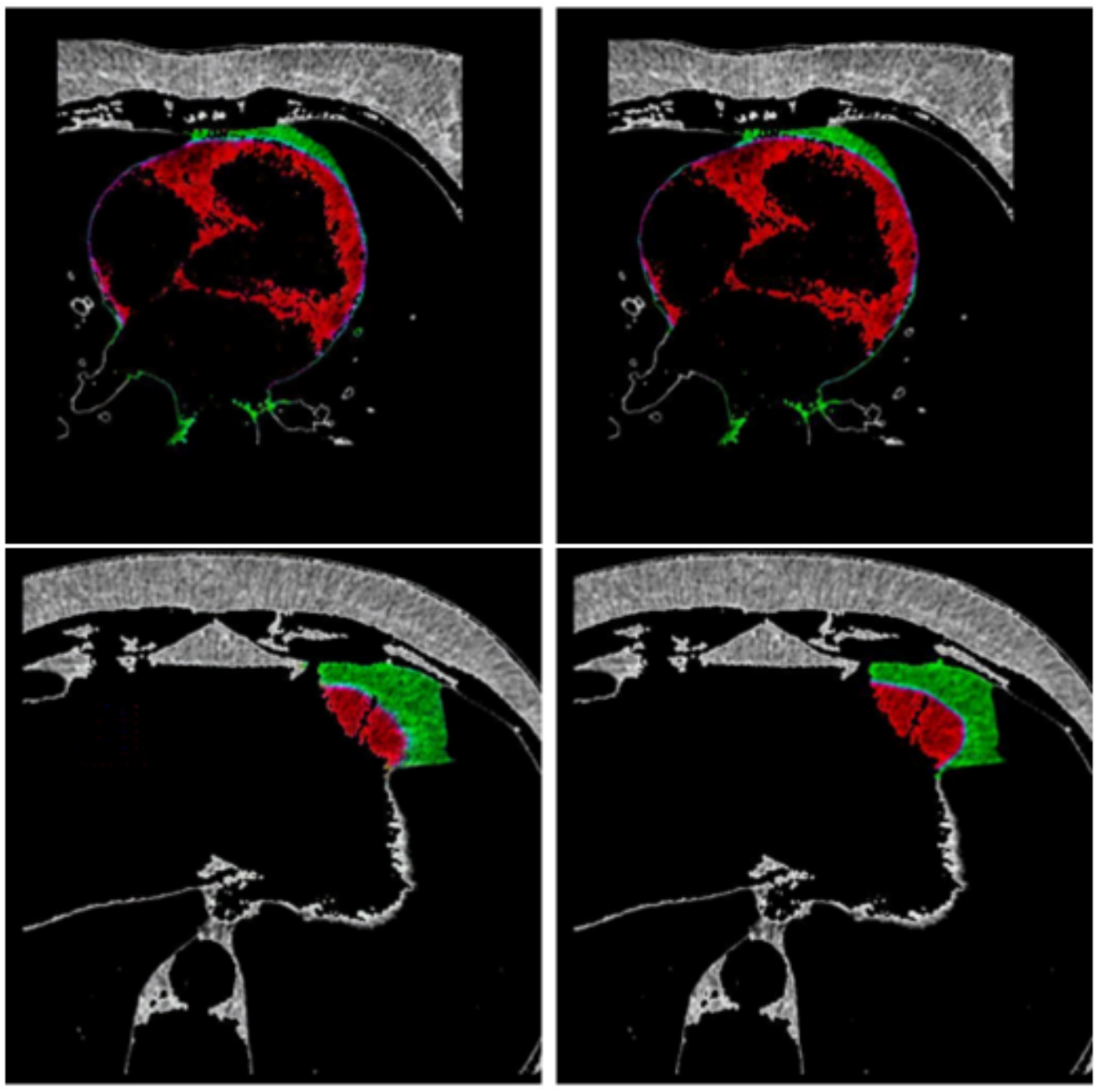

**Fig. 13.** Experiment resizing the images to 256x256. Comparison of segmented images (left) and ground truth images (right).

**Table 3**
Experiment resizing the images to 256x256.

| Class | Acc % | TP % | TN % | $F_1$ % | IoU % | Sec. |
|---|---|---|---|---|---|---|
| Epic. | 98,90 | 99,26 | 98,18 | 99,14 | 98,31 | 01,44 |
| Medias. | 98,97 | 99,23 | 98,36 | 99,22 | 98,46 | 01,44 |
| Peric. | 99,36 | 99,54 | 99,00 | 99,49 | 99,00 | 01,44 |

$$F_1 = \frac{2 \times TP}{2 \times TP + FP + FN} \quad (1)$$

$$IoU = \frac{TP}{TP + FP + FN} \quad (2)$$

## 3. Results

The first experiment carried out aimed to analyze the performance of the trained model with the data in its original format, without any modification and no post-processing performed on the images after segmentation of the images in the test dataset. A few examples of this experiment can be seen in Fig. 13.

Measures calculated to evaluate the performance of this experiment and the time taken to segment an image can be found in Table 3, for the epicardial fat (red), mediastinal fat (green) and pericardium (blue).

The experiment using just the epicardial fat is shown in Fig. 14. The left images are the images generated without post-processing, the ones in the middle are the result after post-processing has been performed (morphological closing), and the left ones are the ground truth.

Table 4 shows the results of this experiment. Therefore, it is now possible to compare the results to the first approach with no post-processing. The post-processing approach slightly reduced the segmentation quality, as the accuracy decreased by 0.05%, the $F_1$ decreased by approximately 0.03% and the IoU decreased by 0.07%. These results indicate that the post-processing, though correcting eventual segmentation flaws (increasing the true positive rate), introduced more noise as well, increasing the total error when compared to the ground truth image.

The third experiment forced the input of the network to deal directly with the size of 512x512 pixels. Fig. 15 shows a few image examples of this experiment.

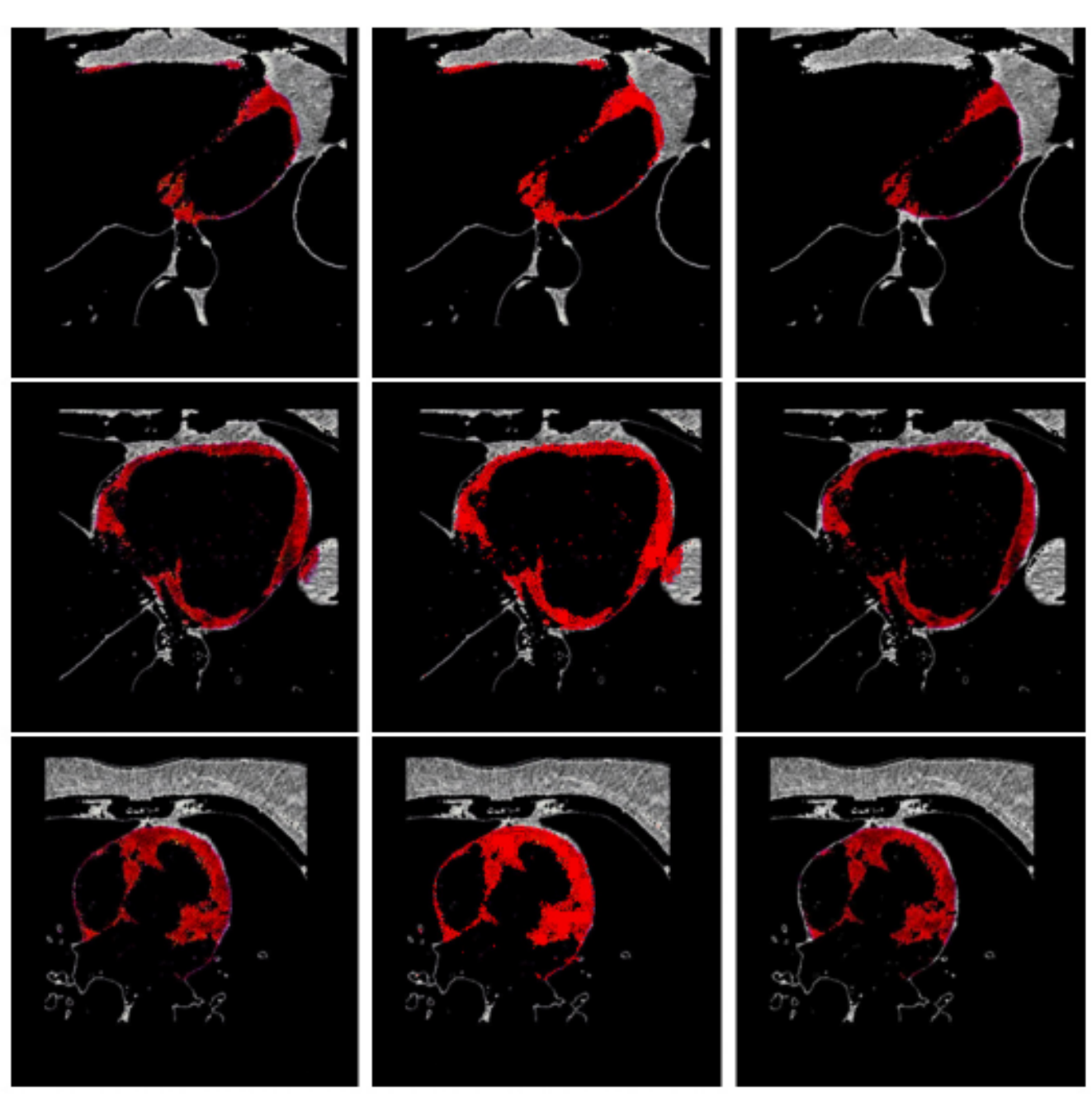

**Fig. 14.** Experiment with the epicardial fat (one fat).

**Table 4**
Experiment with one fat performance measures.

| | Acc % | TP % | TN % | $F_1$ % | IoU % | Sec. |
|---|---|---|---|---|---|---|
| With post-p. | 98,68 | 99,41 | 97,21 | 98,93 | 97,90 | 01,49 |
| Without post-p. | 98,63 | 99,47 | 97,00 | 98,90 | 97,83 | 01,78 |

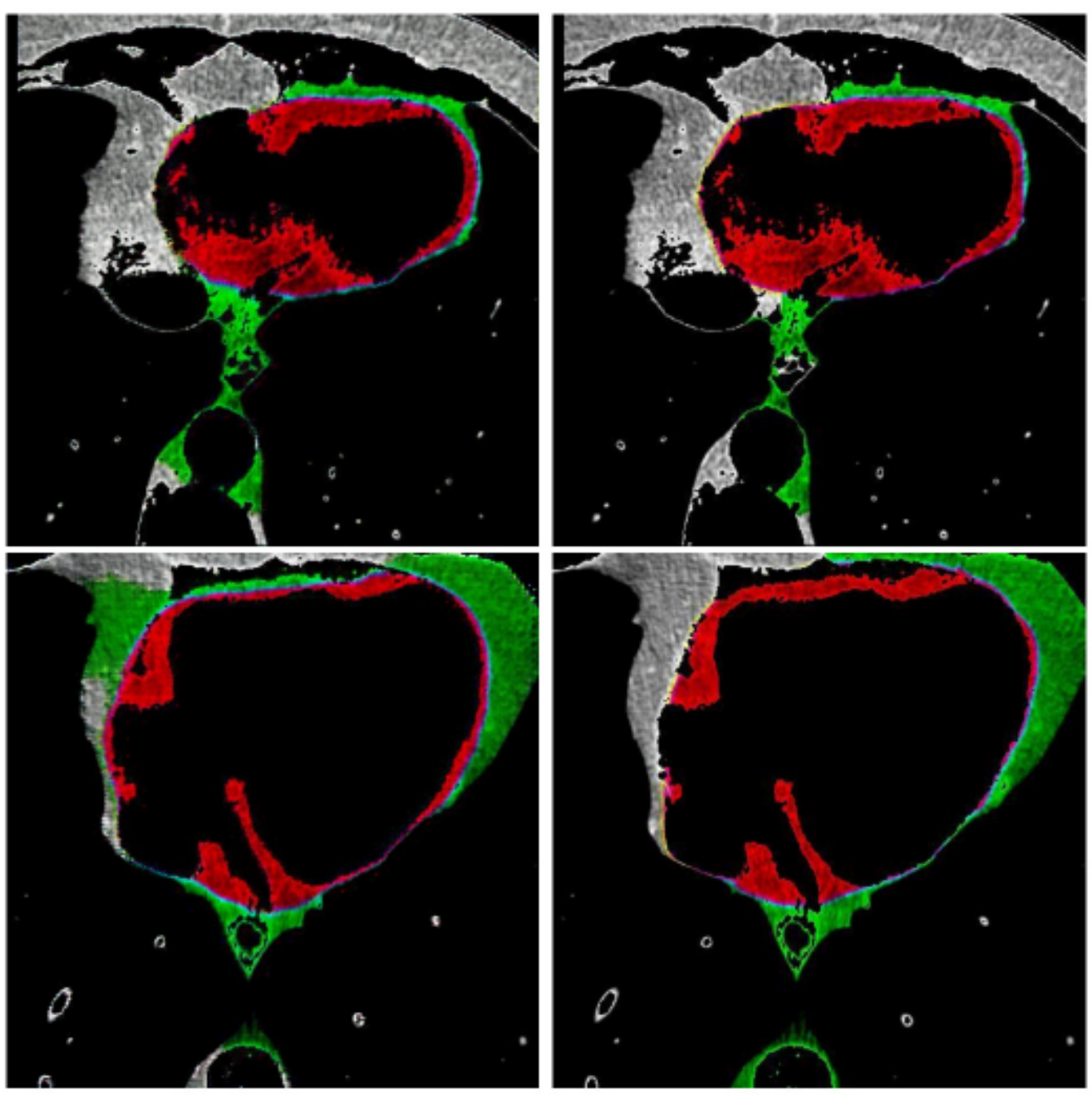

**Fig. 15.** Experiment forcing the input of the network to 512x512. Comparison of segmented images (left) and ground truth images (right).

**Table 5**
Experiment with the original size of the input image (512x512).

| Class | Acc % | TP % | TN % | $F_1$ % | IoU % | Sec. |
|---|---|---|---|---|---|---|
| Epic. | 98,28 | 99,32 | 95,76 | 98,80 | 97,62 | 03,28 |
| Medias. | 98,12 | 99,07 | 95,91 | 98,67 | 97,38 | 03,28 |
| Peric. | 99,21 | 99,55 | 98,42 | 99,42 | 98,86 | 03,28 |

**Table 6**
Experiment with 4 images patches (each 256x256).

| Class | Acc % | TP % | TN % | $F_1$ % | IoU % | Sec. |
|---|---|---|---|---|---|---|
| Epic. | 95,53 | 97,40 | 92,66 | 96,05 | 92,37 | 01,74 |
| Medias. | 96,39 | 97,63 | 94,19 | 96,83 | 93,83 | 01,74 |
| Peric. | 98,45 | 99,15 | 97,37 | 98,45 | 96,95 | 01,74 |

The measures calculated to evaluate the performance of this experiment can be found in Table 5. The observed outcomes reveal that an increase in the quantity of data associated with input images has led to a discernible degradation in segmentation quality. This phenomenon could be attributed to an information surplus necessitating a more intricate network and extended training duration to establish a suitably adept model. This observation could be attributed in part to the standard size of 256x256 that is conventionally employed for the pix2pix network, thus potentially contributing to this effect.

The fourth experiment used the four patches described in the previous section. Each patch contains 256x256 pixels, representing $\frac{1}{4}$ of the original image and some of the results can be seen in Fig. 16. The objective underlying the formulation of this experiment was to assess the network's competence in discerning distinct patterns within individual quadrants. Furthermore, this approach was motivated by the intention to uphold the native dimensions of the network input image, a facet that could potentially yield improvements in segmentation outcomes. Nonetheless, the outcomes of this endeavor did not align with the anticipated outcomes, as elaborated in the subsequent discussion.

The images presented in Fig. 16 depict four patches of the same heart, while Fig. 17 displays the corresponding reconstructed image. This approach also introduces certain drawbacks. First, the segmentation process based on patches can result in decreased capacity to generalize the information, as each input image represents only a fragment of the original heart image and more variation is introduced without providing a representation of the “whole” heart. Additionally, the substantial volume of data generated in this experiment can lead to a greater amount of information, potentially compromising the effectiveness of the segmentation model. More data typically increases challenges in pattern identification, accompanied by extended time demands for the analysis of such augmented datasets.

The performance of the segmentation model of this experiment is highlighted in Table 6. The results obtained are inferior to previous experiments as discussed, indicating that this approach has more disadvantages than benefits for the segmentation model.

## 4. Discussion

This study represents an effort in proposing a fast unified segmentation method that simultaneously addresses both types of cardiac fats, eliminating the need for separate classifiers (one for each type of fat tissue). Moreover, this work introduces the application of the conditional generative adversarial network (cGAN) known as pix2pix for this specific medical segmentation task, thereby presenting novel approaches to harness the capabilities of this network architecture.

In their 2017 work, the authors of the pix2pix framework argued that it may not be the optimal choice for semantic segmentation tasks. However, we demonstrate that this network is capable of yielding satisfactory results in such tasks. Among the experiments conducted in this study, the simplest experiment involving resizing the input images to 256x256 dimensions achieved the highest performance. It achieved an average accuracy of 99.08%, an average true positive rate of 99.34%, and an average $F_1$-score of 99.28%. These results highlight the suitability of the pix2pix network for image segmentation applications.

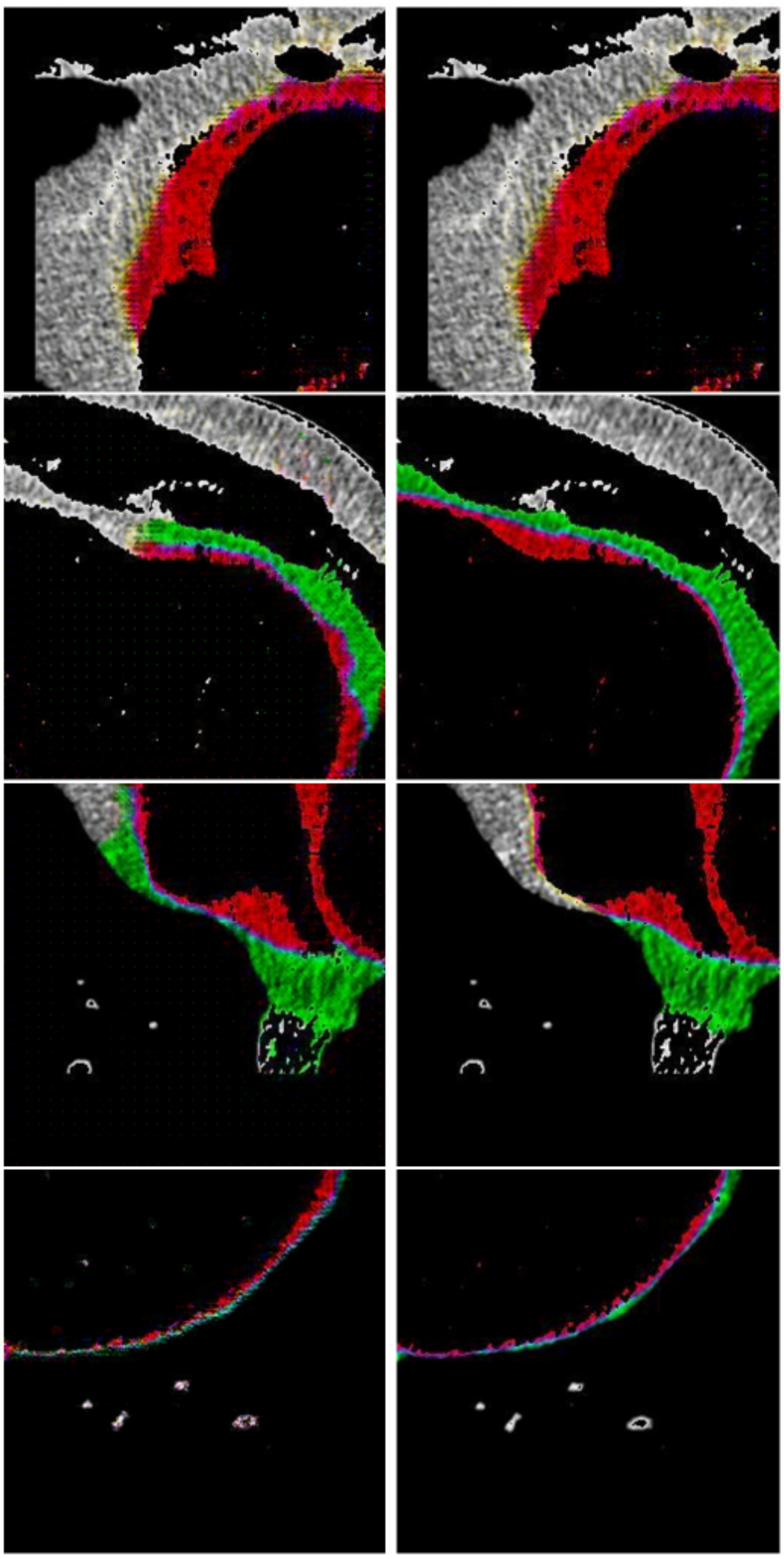

**Fig. 16.** Experiment with the 4 image patches. Comparison of segmented images (left) and ground truth images (right).

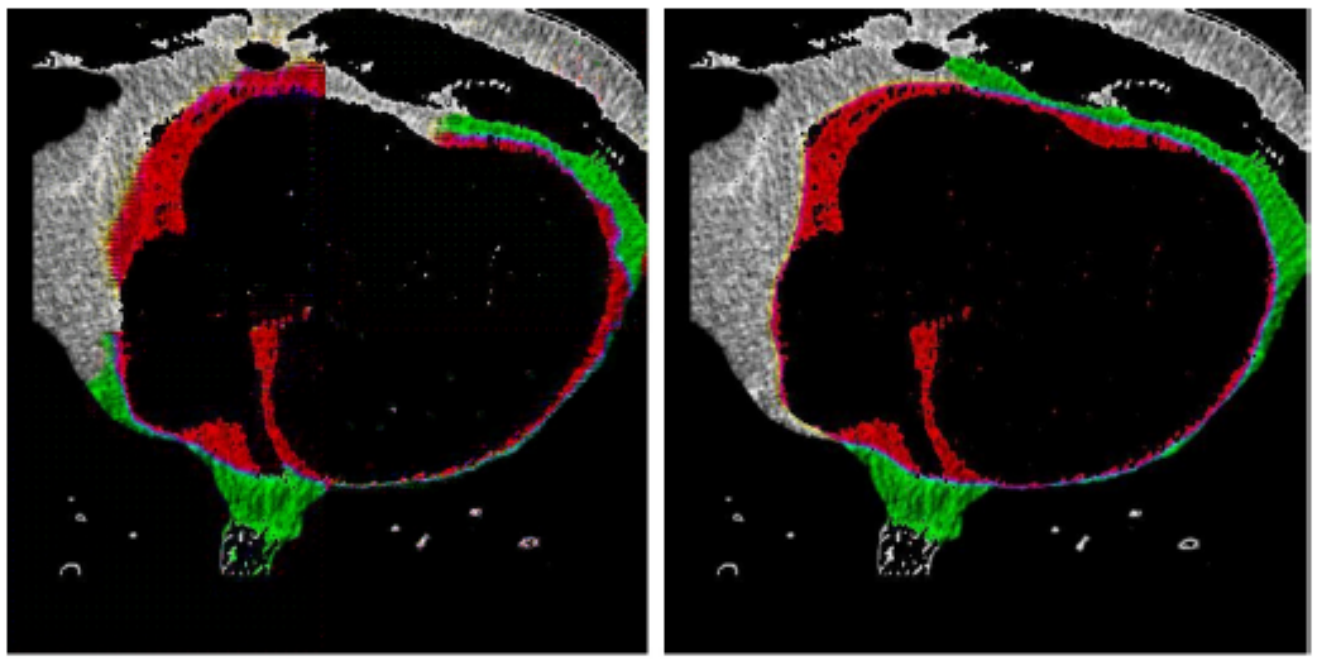

**Fig. 17.** Experiment using the 4 patches. Reconstructed image resulting from the experiment (left) and ground truth (right).

**Table 7**
Comparison of the results obtained in the experiments carried out for the epicardial fat.

| Epicardial | | | | | | |
|---|---|---|---|---|---|---|
| Experiment | Acc % | TP % | TN % | $F_1$ % | IoU % | Time per image (s) |
| Resizing to 256x256 | 98.90 | 99.26 | 98.18 | 99.14 | 98.31 | 01.44 |
| Just the epicardial fat (without post-proces.) | 98.68 | 99.41 | 97.21 | 98.93 | 97.90 | 01.49 |
| Just the epicardial fat (with post-proces.) | 98.63 | 99.47 | 97.00 | 98.90 | 97.83 | 01.78 |
| Using the original size 512x512 | 99.21 | 99.55 | 98.42 | 99.42 | 98.86 | 03.28 |
| 4 image patches (each patch 256x256) | 98.45 | 99.15 | 97.37 | 98.45 | 96.95 | 01.74 |

**Table 8**
Comparison of the results obtained in the experiments carried out for mediastinal fat.

| Mediastinal | | | | | | |
|---|---|---|---|---|---|---|
| Experiment | Acc % | TP % | TN % | $F_1$ % | IoU % | Time per image (s) |
| Resizing to 256x256 | 98.97 | 99.23 | 98.36 | 99.22 | 98.46 | 01.44 |
| Using the original size 512x512 | 98.12 | 99.07 | 95.91 | 98.67 | 97.38 | 03.28 |
| 4 image patches (each patch 256x256) | 96.39 | 97.63 | 94.19 | 96.83 | 93.83 | 01.74 |

**Table 9**
Comparison of the results obtained in the experiments carried out for pericardium.

| Pericardium | | | | | | |
|---|---|---|---|---|---|---|
| Experiment | Acc % | TP % | TN % | $F_1$ % | IoU % | Time per image (s) |
| Resizing to 256x256 | 99.36 | 99.54 | 99.00 | 99.49 | 99.00 | 01.44 |
| Using the original size 512x512 | 98.28 | 99.32 | 95.76 | 98.80 | 97.62 | 03.28 |
| 4 image patches (each patch 256x256) | 95.53 | 97.40 | 92.66 | 96.05 | 92.37 | 01.74 |

The attained processing times in this study are notably remarkable when compared to other approaches reported in the literature, which required several minutes to segment a single image. In contrast, the pix2pix architecture demonstrates a commendable performance, achieving comparable results within seconds. Specifically, the average processing time per image for pix2pix was 1.44 seconds, thereby enabling real-time application of this technology in the day-to-day activities of medical professionals. Some approaches in the literature require several minutes to provide a segmentation result for a single image.

The second experiment focusing solely on epicardial fat segmentation yielded inferior results. It is evident that the application of morphological closing operation effectively eliminated small “holes” present in the images, but it concurrently introduced false positive pixels and noise, thereby increasing the overall error. To potentially enhance future work, one possible solution is to incorporate a noise removal algorithm, such as a Mean Filter. This algorithm replaces the value of a pixel with the median intensity level of its neighboring pixels, which could contribute to improved segmentation accuracy.

For the third and fourth experiments, the ones that considered the original size and 4 patches, respectively, a possible improvement is to reduce the data for training and testing, in order to improve accuracy. Tables 7, 8 and 9 compare the results obtained over all the performed experiments.

We also undertook a paradigm akin to a 3-fold cross-validation procedure. The employed pix2pix framework does not implement a k-fold cross validation internally, and therefore we opted to conduct iterative experiments across the full spectrum of tripartite dataset partitions. This methodological approach enabled us to derive both the arithmetic mean, presented in Table 10, and the corresponding standard deviation, shown in Table 11.

It is important to note that while our approach deviates from the exact conventions of 3-fold cross-validation by virtue of our emphasis on calculating the central tendency, it is anticipated that our outcomes will closely approximate those attained through the natural 3-fold cross-validation technique. In this case we also used the entire 512x512 image (without the patches), resizing it to 256x256 - which obtained the highest rates.

**Table 10**
Average results for the emulated 3-fold cross validation.

| Class | Acc % | TP % | TN % | FN % | F1 | IoU |
|---|---|---|---|---|---|---|
| Epic. | 98.33 | 99.23 | 96.45 | 3.55 | 98.73 | 97.49 |
| Medias. | 97.90 | 99.07 | 95.33 | 4.67 | 98.40 | 96.87 |
| Peric. | 98.96 | 99.53 | 97.87 | 2.13 | 99.17 | 98.37 |

**Table 11**
Standard deviation results for the emulated 3-fold cross validation.

| Class | Acc | TP | TN | FN | F1 | IoU |
|---|---|---|---|---|---|---|
| Epic. | 0.47 | 0.05 | 1.78 | 1.78 | 0.30 | 0.56 |
| Medias. | 0.69 | 0.09 | 2.49 | 2.49 | 0.46 | 0.89 |
| Peric. | 0.30 | 0.02 | 1.02 | 1.02 | 0.20 | 0.39 |

In order to compare the results obtained in this work with others in the literature, the best values reported in each article were used. It is important to point out that the most coherent and fair way to carry out such comparison between the adopted approaches would be with the use of the same execution environment and the same training parameters, from the division of the dataset to the parameters used in the experiments that achieved such results (Table 12).

In order to ensure an fair evaluation, we exclusively considered studies within the literature that employed the identical dataset as suggested by [1]. The methodology presented in this research demonstrated fair accuracy rates across all fat types, effectively competing while also significantly accelerating the segmentation process. Notably, when the f1-score, a more equitable measure of comparison, is taken into account, our proposed methodology surpassed all existing literature contributions.

**Table 12**
Comparison of cardiac fat segmentation results.

| Study | Class | Accuracy (%) | $F_1$ (%) | Time per image (s) |
|---|---|---|---|---|
| Shazad et al. [30] | | – | 89.15 | – |
| Rodrigues et al. (2015) [31] | | 98.50 | 97.9 | 1881 |
| Rodrigues et al. (2016) [1] | | 98.50 | 98.10 | 163 |
| Priya and Sudha [18] | | 98.76 | 98.71 | – |
| Li et al. [23] | Epicardial | – | – | – |
| Kazemi et al. [19] | | 99.50 | 97.80 | – |
| Zhang et al. [20] | | – | 91.19 ± 1.4 | – |
| Albuquerque et al. [21] | | 93.45 | – | 02.01 |
| Li et al. [32] | | 99.58 | – | 4.38 |
| Proposed method | | 98.33 | 98.73 | 01.44 |
| Shazad et al. [30] | | – | – | – |
| Rodrigues et al. (2015) [31] | | 98.40 | – | 1881 |
| Rodrigues et al. (2016) [1] | | 98.40 | – | 163 |
| Priya and Sudha [18] | | 99.23 | 98.26 | – |
| Li et al. [23] | Mediastinal | – | – | – |
| Kazemi et al. [19] | | 99.40 | – | – |
| Zhang et al. [20] | | – | 91.19 ± 1,4 | – |
| Albuquerque et al. [21] | | 93.45 | – | 02.01 |
| Proposed method | | 97.90 | 98.40 | 01.44 |
| Shazad et al. [30] | | – | 89.15 | – |
| Rodrigues et al. (2016) [1] | | 96.40 | – | 163 |
| Priya and Sudha [18] | | 98.20 | 98.52 | – |
| Li et al. [23] | Pericardium | – | – | – |
| Kazemi et al. [19] | | 98.90 | 98.40 | – |
| Zhang et al. [20] | | – | 91.19 ± 1.4 | – |
| Albuquerque et al. [21] | | 93.45 | – | 02.01 |
| Proposed method | | 98.96 | 99.17 | 01.44 |

Reiterating a previously discussed point, it is crucial to highlight the run time performance of our approach, which is currently the fastest in the existing literature. This is an important fact to the real time applications that it implies. Notably, this efficiency stands in stark contrast to the earlier approach presented by [1]. The achieved run times exhibit a very significant enhancement, showcasing a speed improvement of 160 times when compared to the original contribution [1]. Moreover, our methodology run times surpass those of the approach put forth by [21], while concurrently delivering results of high accuracy.

Regrettably, many existing works in the literature lack information regarding processing times, which is a crucial aspect for assessing the applicability of a method. Additionally, authors often do not provide source code or comprehensive instructions for re-implementing their approaches, further complicating the replication process. Even if such resources were available, the computational effort required to re-implement a single approach from other works could extend over several months.

While training times are of some importance, the focus lies on testing times since these are the durations applicable in clinical practice. While it is conceivable to train a model for an extended period, such as a month, to achieve an adequate accuracy, the critical consideration is whether the method can process a single image within a reasonable time frame, such as one second, to facilitate real-time usage. This study takes into account pre-processing times, as well as considering such times in other relevant works.

## 5. Conclusions

In this study, we advocate for the adoption of a conditional generative adversarial network known as pix2pix, which is a deep learning methodology, for the purpose of segmenting cardiac adipose tissue in computed tomography images of the human heart. To determine the most effective approach for addressing this problem, we conducted a series of experiments employing various techniques. Furthermore, we undertook a manual optimization process to refine the parameters of the pix2pix algorithm, ensuring optimal performance in our proposed approach.

Regarding accuracy, which represents the percentage of correctly classified or segmented pixels, the proposed approach exhibited superior performance compared to seven out of eight existing works for epicardial fat segmentation and six out of seven works for mediastinal fat segmentation. Specifically, the approach achieved an average accuracy of 98.33% for epicardial fat and 97.90% for the mediastinal fat. When it comes to the f1-score, our approach outperformed all other works in the literature.

This automation facilitates real-time quantification of cardiac adipose tissue. In comparison to the manual segmentation process conducted by experts, which typically requires approximately 30 minutes for a single image, the proposed approach achieves the same task automatically within seconds on average.

In routine clinical practice, the assessment of adipose tissue related to the human heart, despite its significant relevance, is typically not utilized as a risk factor due to the labor-intensive nature of manual estimation, often resulting in its omission. The introduction of automation techniques allows for efficient and timely estimation of this data, enabling its accessibility to physicians. Moreover, the segmentation of adipose tissues provides visual representation, allowing for the observation of these tissues in two-dimensional or three-dimensional formats.

Moreover, existing literature encompasses studies that investigate the interrelationship between epicardial and mediastinal adipose tissues [27]. Notably, an increase in the quantity of epicardial fat corresponds to an increase in mediastinal fat. Both these adipose tissue depots are considered crucial risk factors associated with various health issues.

Previous investigations, including the current study, have demonstrated the viability of conducting segmentations and quantifications of cardiac fats based on mathematical principles. It is important to underscore that the validation of these findings in clinical practice extends beyond the primary focus of this research. Therefore, future research endeavors could be directed towards collaborative efforts involving clinical experts possessing the requisite knowledge and expertise to effectively address this aspect.

This study centers on two key objectives: achieving efficient processing time for the algorithm without compromising pixel accuracy, and evaluating the feasibility of employing image-to-image architec-tures for image segmentation tasks. Our developed approach facilitates real-time processing, with the algorithm requiring a mere 1 to 2 seconds to quantify the fat content in an image. This speed establishes a new benchmark in the current body of literature. To provide context, if we consider an estimated 50 images per patient, the quantification of the patient's fat content can be completed within a maximum time frame of 2 minutes.

It is worth noting that the utilization of pix2pix parameters posed certain challenges during implementation, necessitating the inclusion of additional filters in the generator output and the discriminator input. Consequently, substantial processing power was required within the execution environment, leading to significantly prolonged training sessions and cases of memory insufficiency. The experiments primarily took place on a system equipped with an Intel quad-core CPU boasting 8 GB of RAM, coupled with an NVIDIA GeForce GTX 1050 TI GPU equipped with 4 GB of dedicated RAM. For further details, including images, source code, and supplementary information, interested readers are directed to the following link: [33].

As a final remark, the dataset utilized in our study encompassed CT images acquired from multiple scanners, predominantly manufactured by two distinct companies, Siemens and Philips. In order to prevent excessive redundancy within the extracted dataset, the incorporating characteristics associated with the scanner model were deliberately omitted.

## Declaration of competing interest

The authors assert the absence of any conflicts of interest. The research received no direct financial backing from external entities. The dataset employed in this study had been previously disseminated in a prior publication by [1] and had undergone scrutiny by ethical review boards. Competing interests: None declared. Funding: None. Ethical approval: Not required.